\documentclass[runningheads]{llncs}
\usepackage[T1]{fontenc}
\usepackage{times}
\usepackage{soul}
\usepackage{url}
\usepackage[hidelinks]{hyperref}
\usepackage[utf8]{inputenc}
\usepackage[small]{caption}
\usepackage{graphicx}
\usepackage{amsmath}
\usepackage{booktabs}
\usepackage[switch]{lineno}

\usepackage{comment}
\usepackage{xcolor}

\usepackage{tikz}

\usepackage{algorithm2e}                                  
\usepackage{algorithmic}                                  

\usepackage{listings}

\begin{document}

\title{Generalized Nested Rollout Policy Adaptation with Limited Repetitions}

\author{
    Tristan Cazenave
}
\authorrunning{Tristan Cazenave}
\institute{
LAMSADE, Université Paris Dauphine - PSL, CNRS, Paris, France
}

\maketitle

\begin{abstract}
Generalized Nested Rollout Policy Adaptation (GNRPA) is a Monte Carlo search algorithm for optimizing a sequence of choices. We propose to improve on GNRPA by avoiding too deterministic policies that find again and again the same sequence of choices. We do so by limiting the number of repetitions of the best sequence found at a given level. Experiments show that it improves the algorithm for three different combinatorial problems: Inverse RNA Folding, the Traveling Salesman Problem with Time Windows and the Weak Schur problem.
\end{abstract}

\section{Introduction}
Monte Carlo Tree Search (MCTS) \cite{Kocsis2006,Coulom2006} has been successfully applied to many games and problems \cite{BrownePWLCRTPSC2012}. It originates from the computer game of Go \cite{Bouzy01} with a method based on simulated annealing \cite{Bruegmann1993MC}. The principle underlying MCTS is to learn the move to play using statistics on random games. In the early times of MCTS, random games were played with a uniform policy. Computer Go program soon used non uniform playout policies, learning the policy with optimization algorithms \cite{Coulom2007}. Playout policies were replaced with neural network evaluations for computer Go with the AlphaGo program \cite{Silver2016MasteringTG}, and then for other games such as Chess and Shogi with the Alpha Zero program \cite{silver2017mastering}. There have been numerous applications of MCTS following the notorious success in Computer Go, ranging from predicting the structure of large protein complexes \cite{bryant2022predicting} to wind farm layout optimization \cite{bai2022wind}.

Nested Monte Carlo Search (NMCS) \cite{CazenaveIJCAI09} is a recursive algorithm which uses lower level playouts to bias its playouts, memorizing the best sequence at each level. After the searches following each possible move have been run, the move of the best sequence at the current level is played. At the lowest level, playouts are performed. They can be uniformly random playouts \cite{CazenaveIJCAI09} or they can be biased using heuristic probabilities for possible moves \cite{portela2018unexpectedly}. Each playout returns the sequence of moves being made and the score of the terminal position. NMCS has given good results on many combinatorial problems such as puzzle solving and single player games \cite{Mehat2010}, the Inverse RNA Folding problem \cite{portela2018unexpectedly} or chemical retrosynthesis \cite{roucairol2023retrosynthesis}.

Nested Rollout Policy Adaptation (NRPA) \cite{Rosin2011}.  combines nested search, memorizing the best sequence of moves found at each level, and the online learning of a playout policy using this sequence. NRPA has world records in Morpion Solitaire and crossword puzzles and has also been applied to many other combinatorial problems such as the Traveling Salesman Problem with Time Windows \cite{cazenave2012tsptw,edelkamp2013algorithm}, 3D Packing with Object Orientation \cite{edelkamp2014monte}, the physical traveling salesman problem \cite{edelkamp2014solving}, the Multiple Sequence Alignment problem \cite{edelkamp2015monte}, Logistics \cite{edelkamp2016monte,Cazenave2021Policy}, Graph Coloring \cite{Cazenave2020Graph}, Vehicle Routing Problems \cite{edelkamp2016monte,Cazenave2020VRP}, Network Traffic Engineering \cite{DangMonteCarlo2021}, Virtual Network Embedding \cite{elkael2022monkey} or the Snake in the Box \cite{dang2023warm}.

Generalized Nested Rollout Policy Adaptation (GNRPA) \cite{Cazenave2020GNRPA} generalizes the way the probability is calculated using a bias. The bias is a heuristic that performs non uniform playouts and using it usually gives much better results than uniform playouts. The use of a bias has been theoretically demonstrated more general than the initialization of the weights. The GNRPA paper also provides a theoretical derivation of the learning of the policy, using a cross entropy loss associated to a softmax. GNRPA has been applied to some difficult problems such as Inverse RNA Folding \cite{Cazenave2020Inverse} and Vehicle Routing Problems \cite{DBLP:journals/corr/abs-2111-06928} with better results than NRPA.

This work presents GNRPA with Limited Repetitions (GNRPALR) a modification to GNRPA that makes it more flexible with regard to the number of iterations at every level. The principle is to stop the iterations at a level when the policy of this level becomes too deterministic. NRPA and GNRPA can waste a lot of time in the last iterations of a level when the policy has become too deterministic as they always replay the same sequence and do not explore alternative sequences anymore in the lower levels. To avoid this behavior we replace the for loop that performs a fixed number of iterations at a level with a while loop that has a threshold on the number of repetitions of the best score at this level.

This paper is organized as follows. The second section describes NRPA, GNRPA and GNRPALR. The third section presents experimental results for three difficult combinatorial problems: Inverse RNA Folding, Traveling Salesman with Time Windows (TSPTW) and Weak Schur Numbers. For these three problems GNRPALR improves much on GNRPA. Moreover the speedups of GNRPALR over GNRPA increase with the search time.

\section{Monte Carlo Search}

This section presents the GNRPA algorithm which is a generalization of the NRPA algorithm to the use of a prior. It also presents the GNRPALR algorithm which is a modification of the GNRPA algorithm to dynamically stop the search at every level.

\subsection{GNRPA}

The Nested Rollout Policy Adaptation (NRPA) \cite{Rosin2011} algorithm is an effective combination of NMCS and the online learning of a playout policy. NRPA holds world records for Morpion Solitaire and crosswords puzzles. 

In NRPA/GNRPA each move is associated to a weight stored in an array called the policy. The goal of these two algorithms is to learn these weights thanks to the best sequences of actions found during the  search. The weights are used in a playout policy that generates good sequences of actions. 

NRPA/GNRPA use nested search. In NRPA/GNRPA, each level takes a policy as input and returns a sequence and its associated score. At any level $>$ 0, the algorithm makes numerous recursive calls to lower levels, adapting the policy each time with the best solution to date. At level 0, NRPA/GNRPA return the sequence of actions generated by the playout function and its associated score.

The playout function sequentially constructs a random sequence of actions biased by the weights of the moves until it reaches a terminal state. It chooses the actions with a probability equal to the application of the softmax function to the weights.

Let $w_{m}$ be the weight associated to a move $m$ in the policy. In NRPA, the probability of choosing move $m$ is defined by: 

$$ p_{m} = \frac{e^{w_{m}}}{\sum_k{e^{w_{k}}}} $$

where $k$ is an element of the set of possible moves, including $m$.

GNRPA \cite{Cazenave2020GNRPA} generalizes the way the probability is calculated using a bias $\beta_{m}$. The probability of choosing move $m$ is:

$$ p_{m} = \frac{e^{w_{m}+\beta_{m}}}{\sum_k{e^{w_{k}+\beta_{k}}}} $$

By taking $\beta_{k} = 0$, we find again the formula for NRPA. 

In NRPA it is possible to initialize the weights according to a heuristic relevant to the problem to solve. In GNRPA, the policy initialization is replaced by the bias. It is sometimes more practical to use $\beta_{k}$ biases than to initialize the weights as the codes for the moves can be different from the codes of the biases.

The algorithm to perform playouts in GNRPA is given in algorithm \ref{PLAYOUT}. The main GNRPA algorithm is given in algorithm \ref{GNRPA}. It calls the adapt algorithm to modify the policy weights so as to reinforce the weights associated to the best sequence of the current level. The policy is passed by reference to the adapt algorithm which is given in algorithm \ref{ADAPT}.

The principle of the adapt function is to increase the weights of the moves of the best sequence of the level and to decrease the weights of all possible moves by an amount proportional to their probabilities of being played.

The definition of $\delta_{bm}$ in the adapt algorithm is:

$$\delta_{bm} = 0 \Leftrightarrow b \neq m$$

$$\delta_{bm} = 1 \Leftrightarrow b = m$$

\begin{algorithm}
\begin{algorithmic}[1]
\STATE{playout ($policy$)}
\begin{ALC@g}
\STATE{$state \leftarrow root$}
\WHILE{true}
\IF{terminal($state$)}
\RETURN{(score ($state$), $sequence(state)$)}
\ENDIF
\STATE{$z$ $\leftarrow$ 0}
\FOR{$m \in$ possible moves for $state$}
\STATE{$o [m] \leftarrow e^{policy[code(m)] + \beta_m}$}
\STATE{$z \leftarrow z + o [m]$}
\ENDFOR
\STATE{choose a $move$ with probability $\frac{o [move]}{z}$}
\STATE{play ($state$, $move$)}
\ENDWHILE
\end{ALC@g}
\end{algorithmic}
\caption{\label{PLAYOUT}The playout algorithm. The moves in the playouts are played with a probability equal to the softmax function applied to the weights plus the bias of the possible moves.}
\end{algorithm}

\begin{algorithm}
\begin{algorithmic}[1]
\STATE{adapt ($policy$, $sequence$)}
\begin{ALC@g}
\STATE{$polp \leftarrow policy$}
\STATE{$state \leftarrow root$}
\FOR{$b \in sequence$}
\STATE{$z \leftarrow 0$}
\FOR{$m \in$ possible moves for $state$}
\STATE{$o [m] \leftarrow e^{policy[code(m)] + \beta_m}$}
\STATE{$z \leftarrow z + o [m]$}
\ENDFOR
\FOR{$m \in$ possible moves for $state$}
\STATE{$p_m \leftarrow \frac{o[m]}{z}$}
\STATE{$polp [code(m)] \leftarrow polp [code(m)] - \alpha(p_m - \delta_{bm})$}
\ENDFOR
\STATE{play ($state$, $b$)}
\ENDFOR
\STATE{$policy \leftarrow polp$}
\end{ALC@g}
\end{algorithmic}
\caption{\label{ADAPT}The adapt algorithm. The moves of the best sequence are reinforced and the probability of playing the possibles moves are subtracted to the weights of the possible moves.}
\end{algorithm}

\begin{algorithm}
\begin{algorithmic}[1]
\STATE{GNRPA ($level$, $policy$)}
\begin{ALC@g}
\IF{level == 0}
\RETURN{playout ($policy$)}
\ELSE
\STATE{$bestScore$ $\leftarrow$ $-\infty$}
\FOR{N iterations}
\STATE{(score,new) $\leftarrow$ GNRPA($level-1$, $policy$)}
\IF{score $\geq$ bestScore}
\STATE{bestScore $\leftarrow$ score}
\STATE{seq $\leftarrow$ new}
\ENDIF
\STATE{$policy$ $\leftarrow$ adapt ($policy$, $seq$)}
\ENDFOR
\RETURN{(bestScore, seq)}
\ENDIF
\end{ALC@g}
\end{algorithmic}
\caption{\label{GNRPA}The GNRPA algorithm. The recursive call and the adapt function are called a fixed number of times at each level.}
\end{algorithm}

\begin{algorithm}
\begin{algorithmic}[1]
\STATE{GNRPALR ($level$, $policy$)}
\begin{ALC@g}
\IF{$level == 0$}
\RETURN{playout ($policy$)}
\ELSE
\STATE{$bestScore$ $\leftarrow$ $-\infty$}
\STATE{$repetitions \leftarrow 0$}
\WHILE{$repetitions \leq R$}
\STATE{$(score,new) \leftarrow$ GNRPALR($level-1$, $policy$)}
\IF{$score == bestScore$}
\STATE{$repetitions \leftarrow repetitions + 1$}
\ENDIF
\IF{$score > bestScore$}
\STATE{$repetitions \leftarrow 0$}
\STATE{$bestScore \leftarrow score$}
\STATE{$seq \leftarrow new$}
\ENDIF
\STATE{$policy \leftarrow$ adapt ($policy$, $seq$)}
\ENDWHILE
\RETURN{$(bestScore, seq)$}
\ENDIF
\end{ALC@g}
\end{algorithmic}
\caption{\label{GNRPALR}The GNRPALR algorithm. The recursive call and the adapt function are called until the score of the best sequence has been sent back by the lower level a fixed number of times.}
\end{algorithm}

\subsection{GNRPALR}

GNRPALR repairs a defect in GNRPA and NRPA. Both algorithms spend a lot of time in the last iterations of a level finding many times the same best sequence at the lower level. They are stuck in a local minima and they do not explore enough to get out of it. In order to avoid this we use a simple measure of how exploratory the policy of the level is. The simple measure is the number of repetitions at the level of the score of the current best sequence of moves. When this number reaches a predefined threshold the recursive calls are stopped and the best sequence is returned. We also experimented with other measures of the exploratory power of the policy such as the entropy of the policy, but the best results were obtained with the number of repetitions. Moreover the number of repetitions is more simple than the entropy of the policy and is easier to understand and to tune.

GNRPALR is described in algorithm \ref{GNRPALR}. It uses the same adapt and playout functions as GNRPA and the structure of the algorithm is similar to GNRPA. The main difference is at line 7 where instead of the for loop that runs a fixed number of iterations in GNRPA, there is a while loop that stops when the algorithm reaches a fixed number of repetitions of the score of the best sequence. The R hyperparameter has to be tuned for each problem. In our experiments the best values range from 0 to 5 repetitions. The number of repetitions is updated at lines 9 to 11. It is reset to 0 at line 13 when a new best sequence is found. The algorithm stops the recursive calls and returns the best score and the best sequence to the level above when the number of times the score of the best sequence has been found at the current level is equal to R.

This is a simple modification to GNRPA that enables to solve problems much faster for long thinking times. Being simple is a quality for an improvement to a search algorithm since it can readily be used by practitioners at a very small development cost and still bring large gains.

\section{Experimental Results}

We now compare GNRPA and GNRPALR for three difficult combinatorial problems: Inverse RNA Folding, TSPTW and Weak Schur Numbers. For each problem we give the evolution of the scores obtained by the algorithms with the logarithm of the search time. Experiments were run on AMD EPYC-Rome processors at 2GHz.

\subsection{The Inverse RNA Folding Problem}

The design of molecules with specific properties is an important topic for health related research. The RNA design problem also named the Inverse RNA Folding problem is a difficult combinatorial problem. This problem is important for scientific fields such as bioengineering, pharmaceutical research, biochemistry, synthetic biology and RNA nanostructures
\cite{portela2018unexpectedly}.

RNA molecules are long molecules composed of four possible nucleotides. Molecules can be represented as strings composed of the four characters A, C, G, U. For RNA molecules of length N, the size of the state space of possible strings is exponential in N. It can be very large for long molecules. The sequence of nucleotides folds back on itself to form what is called its secondary structure. It is possible to find in a polynomial time the folded structure of a given sequence. However, the opposite, which is the Inverse RNA Folding problem, is hard \cite{bonnet2020designing}. 

\begin{figure}
    \centering
    \includegraphics[width=2cm]{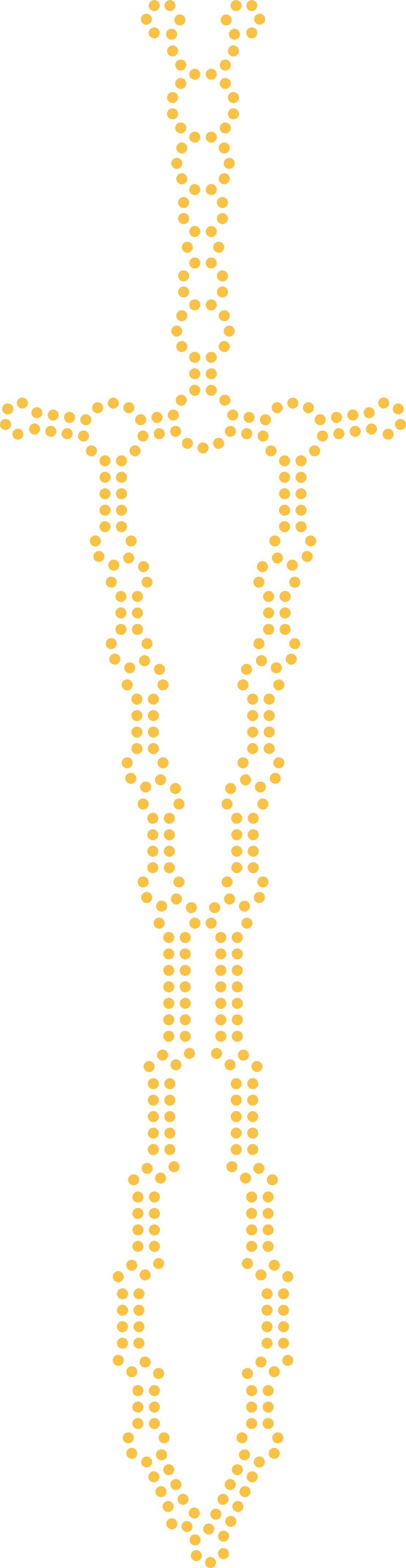}
    \caption{Gladius problem 90 from Eterna100. The associated target structure is:\\
(....)..(....(...(..(.(..(...(((.(((...((((....)))).(((((..(.(((..(.((((..(.((((..(.((((((((.
((((((.(((((.((((.((((.((((...)))).))).)))))).))))).)))))).)))))))..).))))..).))))..).))
)..).))))).)))...(((.(((((.(..(((.(..((((.(..((((.(..(((((((.((((((.(((((.((((((.(((.((((
((....))))))..)))).)))).))))).)))))).)))))))).)..)))).)..)))).)..))).)..))))).((((....))))
...))).)))...)..).)..)...)....)..(....)
}
    \label{gladius}
\end{figure}

We compare Monte Carlo Search algorithms on the Eterna100 benchmark which contains 100 RNA secondary structure puzzles of varying degrees of difficulty. A puzzle consists of a given structure under the dot-bracket notation. This notation defines a structure as a sequence of brackets and dots each representing a base. The matching brackets symbolize the paired bases and the dots the unpaired ones. The puzzle is solved when a sequence of the four nucleotides A, U, G and C, that folds according to the target structure, is found. In some puzzles, the value of certain bases is imposed. Figure \ref{gladius} gives an example of a difficult Eterna100 problem. This is the problem number 90 of the dataset and it is called Gladius.

Human experts have solved the 100 problems of the benchmark. Search algorithms are not yet able to reach this score. The best score so far is 95/100 by NEMO, NEsted MOnte Carlo RNA Puzzle Solver \cite{portela2018unexpectedly}, using NMCS with heuristic playouts, and by GNRPA using the main part of the NEMO prior \cite{Cazenave2020Inverse}.

For our experiments we use a Transformer prior for GNRPA that gives better results than the NEMO prior. To generate the priors we first trained a Transformer policy network \cite{vaswani2017attention} to predict the next nucleotide of the folded sequences using the Rfam dataset \cite{kalvari2021rfam}. We then sampled all the Eterna100 sequences choosing at each step the most probable nucleotide given by the policy. The bias is then computed using a bias of 3.0 if the move is the most probable one. If the move is not the most probable one, with $proba [m]$ the output of the Transformer policy network for move $m$, the bias is: 

$$\beta_m = log (proba [m])$$

The score of a sequence of nucleotide at the end of a playout is computed the same way as NEMO \cite{portela2018unexpectedly} using the ViennaRNA package \cite{lorenz2011viennarna}:

$$
\begin{array}{cc}
     &  
score = \left\{
    \begin{array}{ll}
        \frac{K}{1+\Delta G} & \mbox{if } K>0 \\
        K(1+\Delta G) & \mbox{else}
    \end{array}
\right. \\
& \\
&
    \mbox{with }
    K=1-\frac{BPD}{2*NumTargetPairs}
\end{array}
$$

Where $BPD$ is the number of different pairs between the secondary structure of the sequence and the target structure. $NumTargetPairs$ is the number of pairs in the target structure. $\Delta G$ is the difference between the Minimum Free Energy of the secondary structure and the free energy that the sequence would have in the target structure. 

The objective is to maximize the score function until a value of 1 is obtained, meaning that the problem is solved.

The search for sequences that have a given folding is chaotic. Changing a single nucleotide in a sequence can dramatically change the folding of the sequence and its associated structure. Monte Carlo Search does surprisingly well at finding good sequences in this chaotic search space. The reason could be that it is inherently a sequential decision making algorithm.

\begin{figure}
  \centering
  \includegraphics[width=0.5\textwidth]{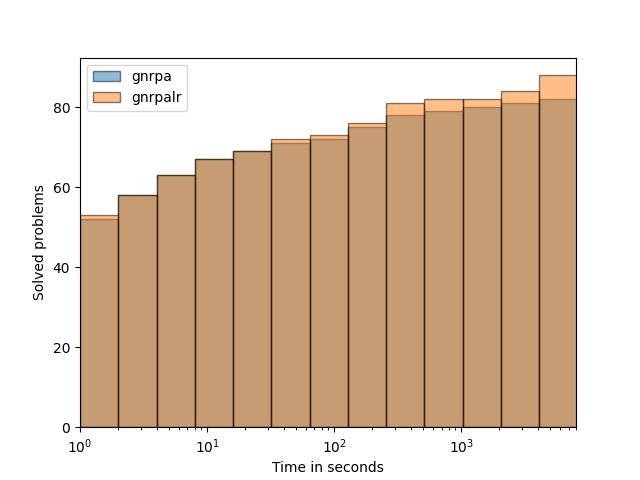}
  \caption{Comparison of GNRPA and GNRPALR for Inverse RNA Folding. The number of repetitions is set to 0 for GNRPALR. GNRPALR is eight times faster than GNRPA. It solves 88 problems in 4,096 seconds when GNRPA solves 82. The relative performance of GNRPALR improves with more search time. The tests are made using the 100 problems of Eterna100.}
  \label{eterna}
\end{figure}

Figure \ref{eterna} gives the comparison between GNRPALR and GNRPA for the 100 problems of Eterna100. The y-axis is the number of problems solved, out of the 100 problems, and the x-axis is the logarithm of the search time. The search times range from 1 second to 4,096 seconds, doubling at every step. We can observe that for short search times the algorithms solve a similar number of problems. For longer search times, and in particular for the 4,096 seconds limit, GNRPALR is much better than GNRPA. The number of repetitions we used for GNRPALR is $R = 0$. This means that for this problem the first repetition is the sign that the while loop should be stopped. The number of possible sequences is huge in the Inverse RNA Folding problems. The probability of finding again the same score for a sequence or the same sequence is quite low. This maybe why finding again the same score is the sign that the policy has become too deterministic.  

\subsection{The Traveling Salesman Problem with Time Windows}

The TSPTW is a practical problem that has everyday applications. NRPA had good results for this problem \cite{cazenave2012tsptw,edelkamp2013algorithm,dang2023warm} as well as for the related Vehicle Routing Problem \cite{edelkamp2016monte,Cazenave2020VRP,Cazenave2021Policy}.

The TSPTW has time constraints represented as time intervals during which cities are to be visited. With Monte Carlo Search, paths with violated constraints can be generated. As presented in \cite{RimmelEvo11} , a new score $score(p)$ of a path $p$ can be defined as follow:
$$
score(p) = - \Omega(p) \times 10^6 -cost(p) 
$$
with, $cost(p)$ the sum of the distances of the path $p$ and $\Omega(p)$ the number of violated constraints. The algorithm uses this evaluation so as to optimize first the number of violated constraints then the sum of the distances between locations of the path.

The prior uses the normalized distance between the current location and the arrival location of the move. Given $max$ the maximum distance between two locations, $min$ the minimum distance between two locations and $d_m$ the distance between the current location and the arrival location, the bias is:

$$\beta_m = 10 \times \frac{d_m - min}{max - min}$$

We experiment with the most difficult instance of the standard test set: the rc204.1 problem \cite{potvin1996vehicle}.

\begin{figure}
  \centering
  \includegraphics[width=0.5\textwidth]{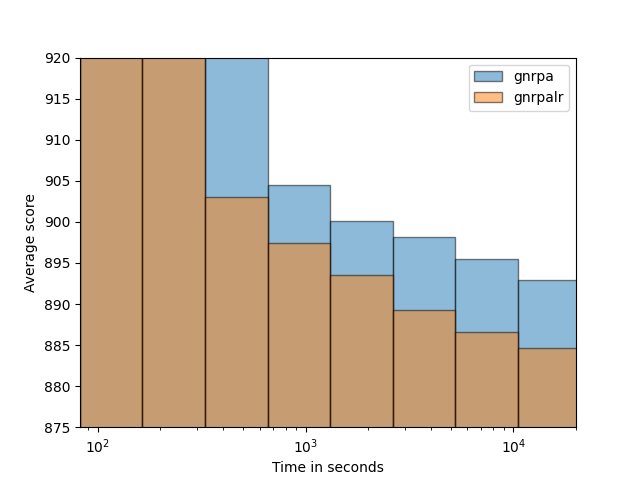}
  \caption{Comparison of GNRPA and GNRPALR for the TSPTW. The number of repetitions is set to 5 for GNRPALR. GNRPALR is much better than GNRPA for this problem. As we can see in the figure, the average score obtained with GNRPA after 10,000 seconds is obtained approximately 8 times slower than with GNRPALR. The averages are calculated over 100 runs of each algorithm with seeds ranging from 1 to 100. The problem solved is rc204.1, the most difficult problem from Solomon test suite for the TSPTW.}
  \label{tsptw}
\end{figure}

Figure \ref{tsptw} gives the comparison between GNRPALR and GNRPA for this instance. GNRPALR finds feasible solutions faster. It finds better makespans faster than GNRPA. It improves the speedup for long search times compared to shorter search times.

\begin{table*}
  \centering
  \caption{Lower bounds for Weak Schur numbers.}
  \label{lowerbound}
  \begin{tabular}{lrrrrrrrrrrrrrr}
 $n$  & 1 & 2 & 3 & 4 & 5 & 6 & 7 & 8 & 9 & 10 & 11 & 12 \\
 $WS(n)$ & 2 & 8 & 23 & 66 & $\geq 196$ & $\geq 646$ & $\geq 2,146$ & $\geq 6,976$ & $\geq 22,536$ & $\geq 71,256$ & $\geq 243,794$ & $\geq 815,314$ \\
 \end{tabular}
\end{table*}

\subsection{The Weak Schur Problem}

The Weak Schur problem is to find a partition of consecutive numbers that contains as many consecutive numbers as possible, where a partition must not contain a number that is the sum of two previous numbers in the same partition.

The score of a terminal partition is the last number that was added to the partition before the next number could not be placed. The goal is to find partitions with high scores. These scores are lower bounds on the exact Weak Schur numbers.

An optimal partition of size 3 is for example:

\begin{verbatim}
1 2 4 8 11 22
3 5 6 7 19 21 23
9 10 12 13 14 15 16 17 18 20    
\end{verbatim}

And thus $WS(3) = 23$. The current records for the Weak Schur problem are given in table \ref{lowerbound} from \cite{ageron2021new}.

When possible, it is often a good move to put the next number in the same partition as the previous number. We use the same selective policy as in \cite{cazenave2016selective} which follows this heuristic. If it is legal to put the next number in the same partition as the previous number then it is the only legal move considered. Otherwise all legal moves are considered. The code of a move for the Weak Schur problem takes as input the partition of the move, the integer to assign and the previous number in the partition.

\begin{figure}
  \centering
  \includegraphics[width=0.5\textwidth]{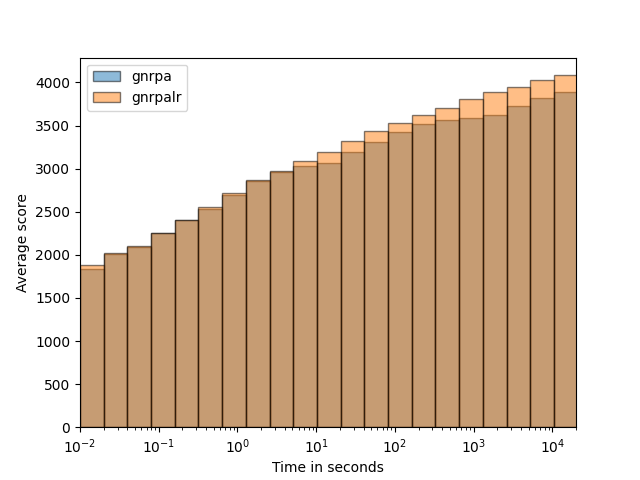}
  \caption{Comparison of GNRPA and GNRPALR for the Weak Schur problem of dimension 8. The number of repetitions is set to 0 for GNRPALR. The averages are calculated over 100 runs of each algorithm with seeds ranging from 1 to 100. The improvement of GNRPALR over GNRPA is greater for long search times. The average score obtained with GNRPA after 10,000 seconds is obtained approximately 8 times slower than with GNRPALR.}
  \label{ws}
\end{figure}

The comparison between GNRPA and GNRPALR for dimension 8 is given in Figure \ref{ws}. We can observe that the difference in average score increases with the search time. It means that the lower bounds found by GNRPALR for long search times take much longer to be found by GNRPA, whereas the algorithms have similar results for short search times.

\section{Conclusion}

Enforcing a limited number of repetitions for GNRPA at the different levels of the nested searches avoids deterministic policies. It enables to search longer when discovering new better sequences and to stop search when the algorithm finds again and again the same best sequence. It speeds up GNRPA for three difficult combinatorial problems with thinking times up to 10,000 seconds for TSPTW and Weak Schur and 4,096 seconds for Inverse RNA Folding. The speedups are greater when the search time is longer. For Inverse RNA Folding, TSPTW and Weak Schur the speedup is approximately eight fold when searching during the longest tested time.

Future work will experiment with GNRPALR for other difficult combinatorial problems. It could also be interesting to tailor the $R$ hyperparameter to the level as it costs less to have a great $R$ at a level than in all levels. For example having twice the number of playouts at every level costs $2^L$ more times for a search at level $L$, when having twice the number of playouts at the lowest level only costs twice the time. If the policy is less deterministic due to a smaller $R$ in the upper levels, the lowest level is naturally more exploratory. This is close to the idea of Stabilized NRPA \cite{Cazenave2020Stabilized} which performs more playouts at the lowest level while keeping the same number of adapt. Stabilized NRPA has already proven beneficial for SameGame, TSPTW and Expression Discovery .


\bibliographystyle{plain}
\bibliography{main}

\end{document}